\documentclass{article}

\usepackage{microtype}
\usepackage{graphicx}
\usepackage{booktabs} % for professional tables
\usepackage[table]{xcolor}
\usepackage{bbding} %特殊符号
\usepackage{hyperref}
\usepackage{arydshln}

\usepackage[accepted]{icml2024}

\usepackage{amsmath}
\usepackage{amssymb}
\usepackage{mathtools}
\usepackage{amsthm}
\usepackage{subcaption}
\usepackage{enumitem}
\usepackage{multirow}
\usepackage{multicol}

\usepackage[capitalize,noabbrev]{cleveref}

\theoremstyle{plain}

\theoremstyle{definition}

\theoremstyle{remark}

\usepackage[textsize=tiny]{todonotes}

\icmltitlerunning{Image Fusion via Vision-Language Model}

\newcommand{\bfsection}[1]{\vspace*{0.00cm}\noindent\textbf{#1.}}
\graphicspath{{Figure/}}
\usepackage{caption}
\definecolor{secondcolor}{RGB}{220,230,240}
\definecolor{firstcolor}{RGB}{241,220,219}

\begin{document}
    
    \twocolumn[
    \icmltitle{Image Fusion via Vision-Language Model}
    \begin{icmlauthorlist}
        \icmlauthor{Zixiang Zhao}{xjtu,eth}
        \icmlauthor{Lilun Deng}{xjtu}
        \icmlauthor{Haowen Bai}{xjtu}
        \icmlauthor{Yukun Cui}{xjtu}
        \icmlauthor{Zhipeng Zhang}{eth,nwpu}
        \icmlauthor{Yulun Zhang}{sjtu}\\
        \icmlauthor{Haotong Qin}{eth}
        \icmlauthor{Dongdong Chen}{hw}
        \icmlauthor{Jiangshe Zhang}{xjtu}
        \icmlauthor{Peng Wang}{nwpu}
        \icmlauthor{Luc Van Gool}{eth,Leuven,insait}
    \end{icmlauthorlist}
    \icmlaffiliation{xjtu}{Xi'an Jiaotong University, China}
    \icmlaffiliation{eth}{ETH Z\"urich, Switzerland}
    \icmlaffiliation{nwpu}{Northwestern Polytechnical University, China}
    \icmlaffiliation{sjtu}{MoE Key Lab of Artificial Intelligence, AI Institute, Shanghai Jiao Tong University, China}
    \icmlaffiliation{hw}{Heriot-Watt University, United Kingdom}
    \icmlaffiliation{Leuven}{KU Leuven, Belgium}
    \icmlaffiliation{insait}{INSAIT, Bulgaria}
    \icmlcorrespondingauthor{Yulun Zhang}{yulun100@gmail.com}
    \icmlcorrespondingauthor{Dongdong Chen}{d.chen@hw.ac.uk}
    \icmlkeywords{Machine Learning, ICML}
    \vskip 0.3in
    ]
    
    \printAffiliationsAndNotice{}  % leave blank if no need to mention equal contribution
    
    \begin{abstract}
        Image fusion integrates essential information from multiple images into a single composite, enhancing structures, textures, and refining imperfections. Existing methods predominantly focus on pixel-level and semantic visual features for recognition, but often overlook the deeper text-level semantic information beyond vision. Therefore, we introduce a novel fusion paradigm named \textit{image \textbf{F}usion via v\textbf{I}sion-\textbf{L}anguage \textbf{M}odel} (\textbf{FILM}), for the first time, utilizing explicit textual information from source images to guide the fusion process.
        Specifically, FILM generates semantic prompts from images and inputs them into ChatGPT for comprehensive textual descriptions. These descriptions are fused within the textual domain and guide the visual information fusion, enhancing feature extraction and contextual understanding, directed by textual semantic information via cross-attention.
        FILM has shown promising results in four image fusion tasks: infrared-visible, medical, multi-exposure, and multi-focus image fusion. We also propose a vision-language dataset containing ChatGPT-generated paragraph descriptions for the eight image fusion datasets across four fusion tasks, facilitating future research in vision-language model-based image fusion. Code and dataset are available at \url{https://github.com/Zhaozixiang1228/IF-FILM}.
    \end{abstract}

    \section{Introduction}\label{sec:1}
    \begin{figure}[t]
        \centering
        \includegraphics[width=\linewidth]{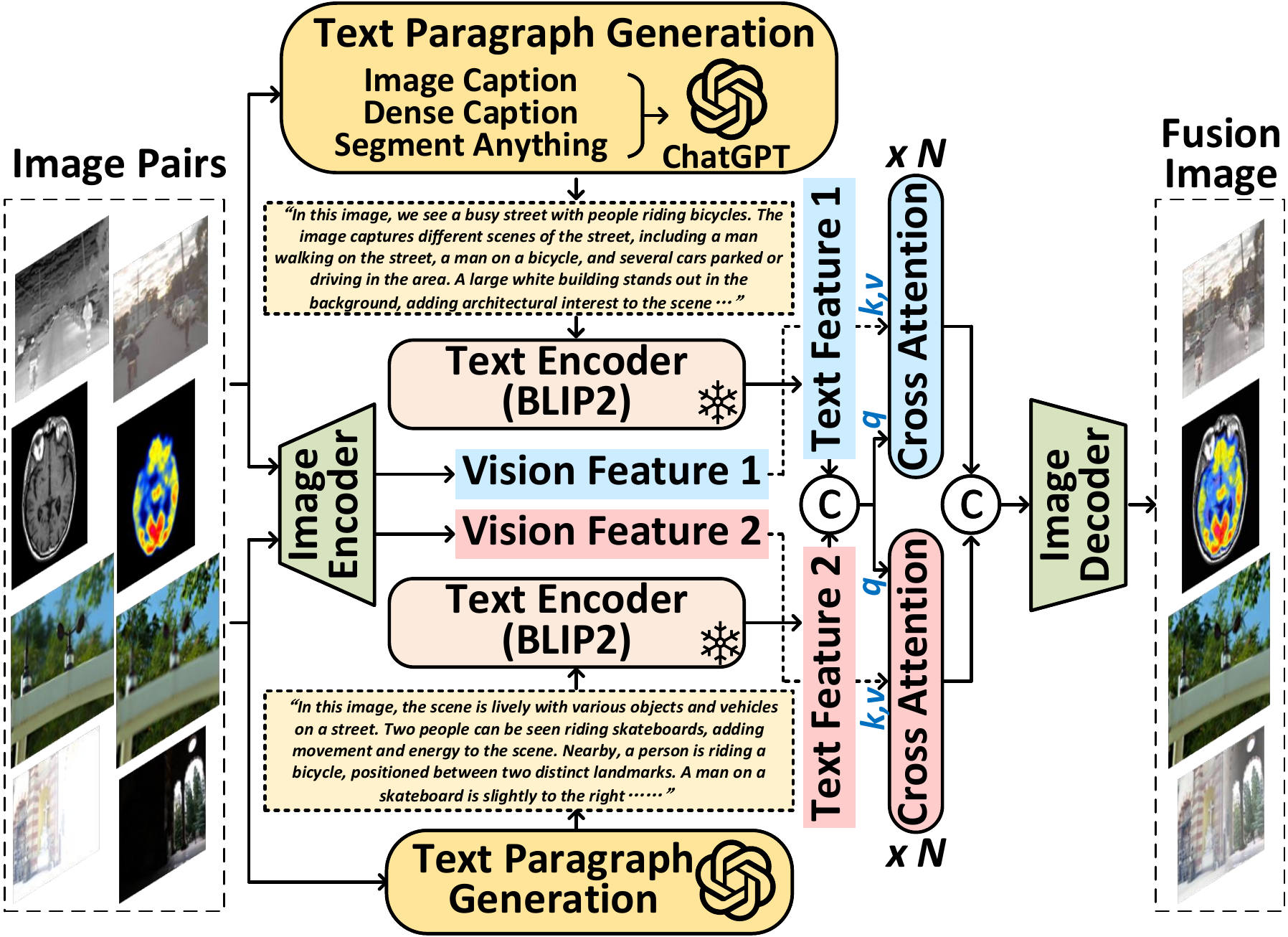}
        \caption{Workflow for our FILM. Input images are first processed to create prompts for ChatGPT \cite{chatgpt}, which then generate detailed textual descriptions. These descriptions help to get fused textual features via the frozen BLIP2 \cite{DBLP:conf/icml/0008LSH23} model. Then, these textual features are fused and guide the extraction and fusion of visual features via cross-attention, enhancing contextual understanding with text-based semantic information. Finally, the fusion image is output by the image decoder.}
        \label{fig:Workflow}
    \end{figure}
    
    Image fusion \cite{Zhao_2023_CVPR,Zhao_2023_ICCV,DBLP:conf/cvpr/LiuFHWLZL22,zhang2021deep}, standing as a critical technique in computer vision, combines information from multiple source images to create a single image that is more informative and suitable for human or machine perception.
    The realm of image fusion encompasses several sub-tasks, each addressing unique challenges and applications. Representatively, infrared-visible image fusion combines infrared and visible images, enhancing object representation under varied illumination conditions \cite{Zhao_2023_CVPR,Zhao_2023_ICCV}. Medical image fusion integrates different modalities of clinical images such as MRI and CT scans, offering a more comprehensive view for diagnosis and treatment planning \cite{DBLP:journals/inffus/JamesD14,DBLP:journals/inffus/Xu021,10190200}. Multi-exposure image fusion merges images taken with different exposure settings to capture a wider range of luminance, which is crucial in high dynamic range imaging \cite{DBLP:journals/tip/MaDZFW20,DBLP:journals/tip/MaLYWMZ17}. Lastly, multi-focus image fusion merges images focused on different planes to produce a uniformly sharp image, invaluable in microscopy and macro photography \cite{DBLP:journals/pami/0002D21,zhang2021deep}.
    
    Despite its widespread application, the current state of image fusion is marred by a notable limitation: an {over-reliance} on visual features. The prevalent methodologies in this field predominantly center around the vision feature extraction, alignment, fusion, and reconstruction, prioritizing aspects like texture, contrast, highlight information, and pixel registration \cite{Zhao_2023_CVPR,Liang2022ECCV}. This consequently neglects the deeper, semantic layers of information that images inherently possess.
    Approaches that integrate downstream pattern recognition tasks like semantic segmentation \cite{DBLP:journals/ieeejas/TangDMHM22,DBLP:journals/inffus/TangYM22,Liu_2023_ICCV} and object detection \cite{DBLP:conf/cvpr/LiuFHWLZL22,DBLP:conf/mm/SunCZH22,DBLP:conf/cvpr/ZhaoXZHL23}, although progressive, still fall short as they remain concentrate on the superficial semantics derived from visual pixel level cues rather than the deeper, more nuanced textual information that images can convey. Therefore, how to better utilize the deeper-semantic features that go beyond visual information in images, becomes a breakthrough point that urgently needs to be addressed.
    
    With the development of large language models \cite{chatgpt}, some work \cite{zhang2023gpt4roi,zhu2023minigpt,radford2021learning,OpenAI_GPT4_2023} attempts to utilize Vision Language Model (VLM) for information fusion and alignment as supplementary.
    These models, which include notable architectures like CLIP \cite{radford2021learning} and GPT4 \cite{OpenAI_GPT4_2023}, demonstrate remarkable proficiency in understanding and generating content that synergizes visual and textual components. They not only tap into the knowledge capabilities of the large language model, but also align and fuse with visual information \cite{DBLP:conf/icml/0008LSH23}. This synergy has the potential to significantly enhance image fusion processes, offering a pathway to incorporate deeper semantic understanding guided by language, thereby enabling a more comprehensive and contextually rich fusion process. For instance, when describing two multi-modal images from the same scene, the descriptions should focus on the response characteristics unique to each modality; for descriptions of multi-focus images, the text should pay greater attention to the areas of perfect imaging that align with their focal points. Thus, we can extract textual descriptions from images based on the large vision-language model and, after integrating descriptions on the textual feature level, we then use fused text features to guide the extraction and fusion of features at the image and vision level.

    Therefore, in this paper, we propose a novel fusion algorithm called \textit{Image \textbf{F}usion via V\textbf{I}sion-\textbf{L}anguage \textbf{M}odel} (\textbf{FILM}). This approach integrates the capabilities of VLMs into the image fusion process, for the first time, leveraging the semantic understanding derived from textual data to guide and enhance the fusion of visual features. Our methodology comprises three components: text feature fusion, language-guided vision feature fusion, and vision feature decoding. The workflow of these components is depicted in \cref{fig:Workflow}. Our contributions are summarized as follows:
    \begin{itemize}[itemsep=0.1cm,topsep=0pt,parsep=0pt]
        \item We propose a novel paradigm for image fusion. To our knowledge, this is the first instance of incorporating explicit (\emph{language model derived}) textual guidance into image fusion algorithms. This approach aids in the deeper understanding of text-level semantic information, facilitating the extraction and fusion of strengths from each source image.
        \item Our model has achieved satisfactory results in infrared-visible, medical, multi-exposure, and multi-focus image fusion tasks, demonstrating its effectiveness across various application scenarios.
        \item We introduce a series of vision-language benchmark datasets for image fusion covering eight fashion datasets across four fusion tasks. These datasets comprise manually refined prompts tailored for the ChatGPT model, alongside paired textual descriptions generated by ChatGPT, which are designed to facilitate subsequent research in image fusion using vision-language models.
    \end{itemize}
    
    \section{Related Work}\label{sec:2}
    In this section, we will review the image fusion algorithms in the era of deep learning~(DL) and introduce the key technology used in our paper: the Vision-Language Model (VLM).
    
    \bfsection{DL-Based Image Fusion}
    In the era of deep learning, neural networks are often used for source feature extraction, feature fusion, and fused image reconstruction \cite{Zhao_2023_CVPR,zhang2021deep}.
    \textbf{(a)} In \textit{multi-modal image fusion}, since there is no ground truth available, it inherently belongs to an unsupervised task. The fusion methods can be divided into generative and discriminative categories \cite{10088423}. Generative algorithms model the latent space manifold through the generative adversarial network (GAN) \cite{ma2019fusiongan,DBLP:conf/cvpr/LiuFHWLZL22} or denoising diffusion model \cite{Zhao_2023_ICCV}, making the distribution gap between source and fused images as close as possible. On the other hand, discriminative models, based on regression ideas, use the model-driven \cite{DBLP:conf/cvpr/Xu0ZSL021,DBLP:journals/tcsv/ZhaoXZLZL22,li2023lrrnet,DBLP:conf/cvpr/ZhaoZXLP22} or data-driven \cite{zhaoijcai2020,Zhao_2023_CVPR,li2018densefuse} auto-encoder structures to learn the source-fusion images mapping. Additionally, downstream cross-modal pattern recognition tasks, such as object detection \cite{DBLP:conf/cvpr/LiuFHWLZL22,DBLP:conf/mm/SunCZH22,DBLP:conf/cvpr/ZhaoXZHL23} and semantic segmentation \cite{DBLP:journals/ieeejas/TangDMHM22,DBLP:journals/inffus/TangYM22,Liu_2023_ICCV}, are employed to make the fused images highlight features and regions containing vision-based semantic information \cite{DBLP:conf/cvpr/ZhaoXZHL23}.
    \textbf{(b)} In \textit{digital image fusion}, supervised fusion algorithms often obtain a mapping from imperfect source images to perfect ground truth by predicting decision maps or reconstructing images \cite{DBLP:journals/pami/0002D21,liu2017multi,xiao2021dtmnet,xiao2020global,ma2020alpha,yin2021automatic}. For issues where perfect training ground truth, like normally-exposed and all-focus images, are hard to obtain, unsupervised algorithms often reconstruct fused images based on CNN \cite{prabhakar2017deepfuse,han2022multi,DBLP:journals/tip/GaoDXXD22,bouzos2023convolutional,guan2023mutual,deng2021deep}, Transformer \cite{qu2022transmef,guan2023mutual}, or GAN \cite{yin2021two,cai2018learning,xu2020mef,DBLP:journals/tmm/GuoNCZMH19}. The feature substitution or fusion, with the help of no-reference quality metrics \cite{xu2020towards,liu2023holoco}, usually occurrence in image domains, frequency domains, or feature spaces \cite{ma2020alpha,wang2022self,wang2023multi}.
    \textbf{(c)} Furthermore, registration-based methods focus on solving the misalignment issue in multi-source image inputs, reducing artifacts in the fused images \cite{DBLP:conf/ijcai/WangLFL22,DBLP:conf/mm/JiangZ0L22,DBLP:conf/cvpr/Xu0YLL22}. Meanwhile, unified frameworks explore the meta-information in different fusion tasks, investigating the mutual promotion effects and alleviating the issues of lacking paired training data and absence of ground truth \cite{9151265,Liang2022ECCV}.
    
    \bfsection{Vision-Language Model}
    Recently, visual language multi-modal learning \cite{radford2021learning,DBLP:conf/icml/0001LXH22,xu2015show,brooks2023instructpix2pix,qin2024accurate,huang2024billm} has become a hot research topic. In particular, vision-language models  \cite{zhu2023minigpt,wang2021simvlm,DBLP:conf/icml/0001LXH22,OpenAI_GPT4_2023,zhang2023gpt4roi} such as BLIP \cite{DBLP:conf/icml/0001LXH22,DBLP:conf/icml/0008LSH23}, DALL-E \cite{ramesh2021zero,ramesh2022hierarchical}, and GPT4 \cite{OpenAI_GPT4_2023} have shown powerful performance in several downstream tasks. BLIP demonstrates powerful knowledge prompt capabilities in bridging between frozen visual feature pre-trained encoders and frozen large language models. GPT4 also shows strong general performance based on visual language pre-training. With the help of these large models \cite{touvron2023llama,OpenAI_GPT4_2023}, a lot of studies \cite{zhang2023gpt4roi,zhu2023minigpt} in image captioning have turned into guiding the large models to provide detailed descriptions of images in the form of natural language. These large models provide external common knowledge for image caption. The key details information from the image such as dense caption, can provide a strong explicit prompt. It allows the image to be presented in a descriptive form that covers the key information. Inspired by this, we aim to introduce a vision-language model to image fusion so that text can guide image fusion in an effective and intuitive way.
    
    \bfsection{Comparison with Existing Approaches}
    The most closely related approaches to our method are the ones that use pattern recognition tasks to provide guidance through visual semantic information \cite{DBLP:journals/inffus/TangYM22,Liu_2023_ICCV,DBLP:conf/cvpr/LiuFHWLZL22,DBLP:conf/cvpr/ZhaoXZHL23}. In contrast to such methods, our approach transcends the limitations of visual semantic information by utilizing deeper-level textual semantic information, guiding the feature extraction and selection in fusion tasks through language and textual features. The integration of VLM in image fusion tasks promises a transformative shift, enabling a more holistic understanding of images through the combined perspectives of both visual perception and textual context, thereby paving the way for more sophisticated and application-specific fusion techniques.
    \begin{figure*}[t]
        \centering
        \includegraphics[width=\linewidth]{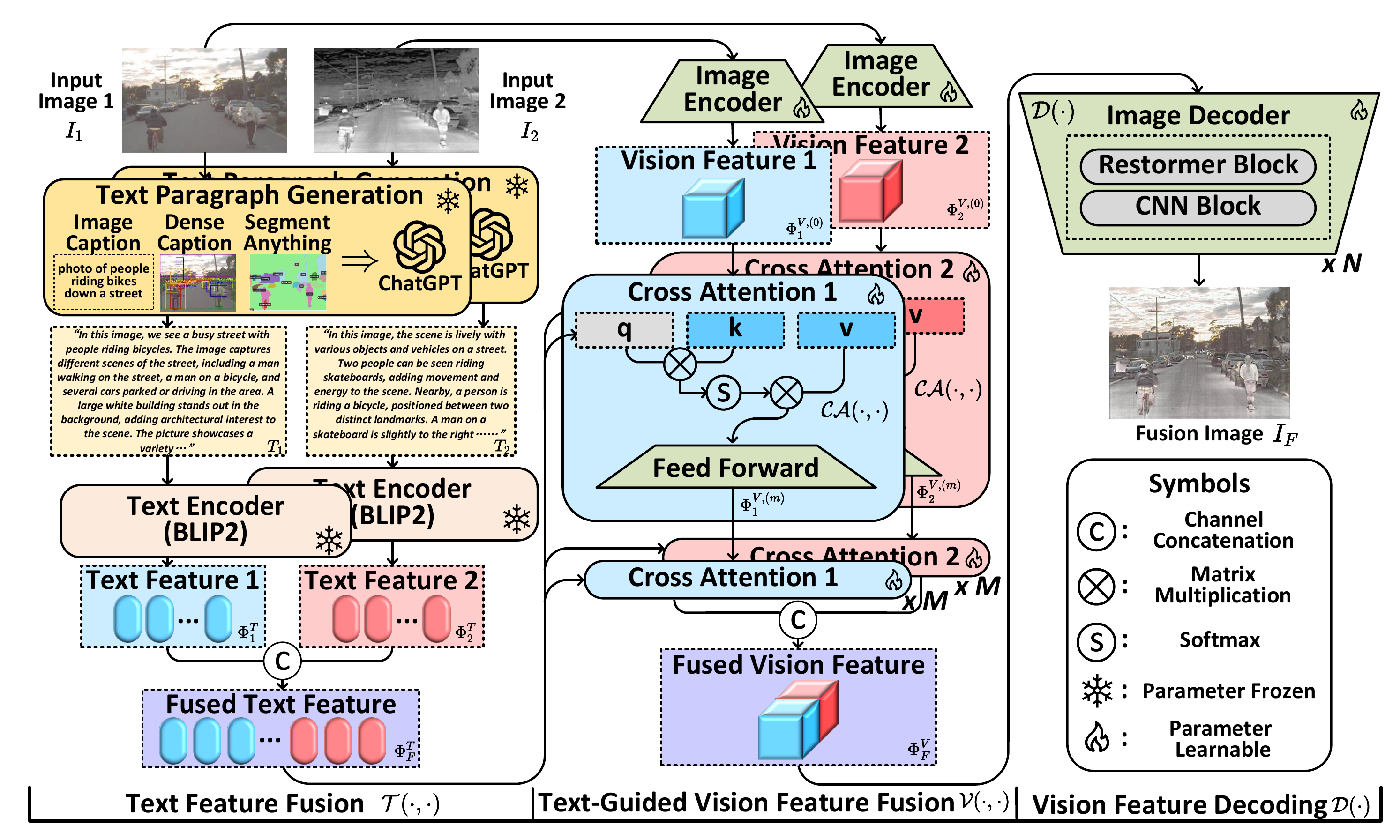}
        \caption{Workflow for our FILM, which encompasses three components: text paragraph generation and text feature fusion, language-guided vision feature fusion via cross attention and vision feature decoding, corresponding to the first, second, and third columns.}
        \label{fig:Workflow2}
    \end{figure*}
    
    \begin{figure*}[t]
        \centering
        \includegraphics[width=\linewidth]{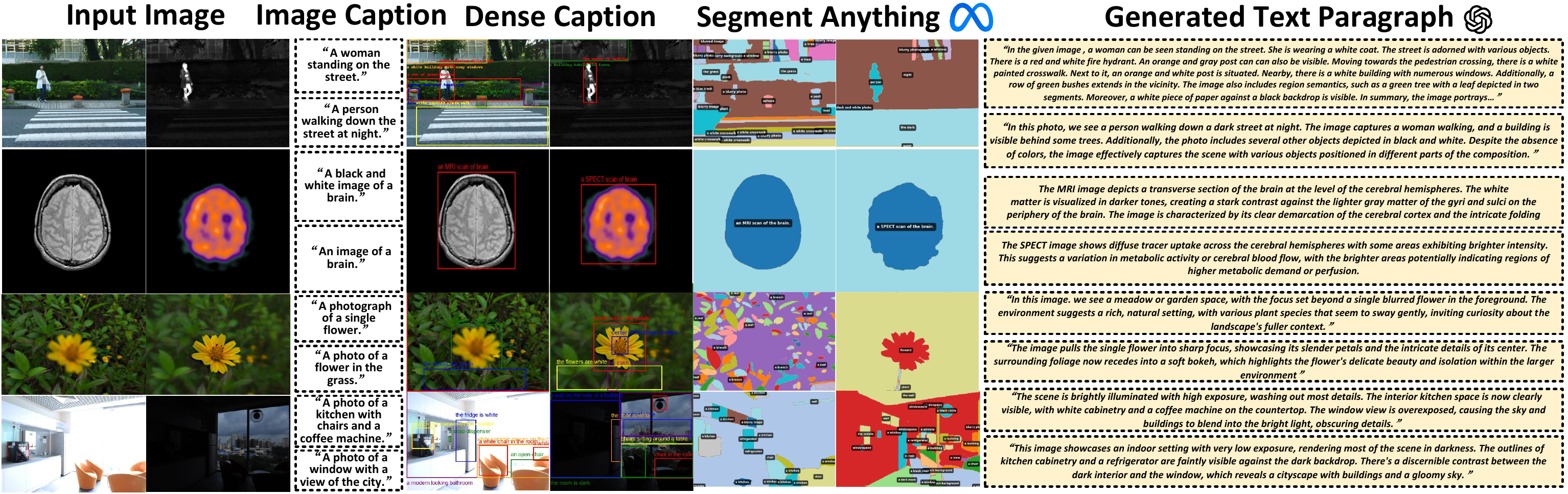}
        \caption{Visualization of the VLF dataset creation process and representative data displays.}
        \label{fig:WorkflowDataset}
    \end{figure*}
    
    \section{Method}
    In this paper, we denote the input pairs of images, which may be a pair of infrared-visible, medical, multi-exposure, or multi-focus images, as $I_1$ and $I_2$. The algorithm ultimately outputs a fused image, represented as $F$. In this section, we will provide a comprehensive description of our FILM algorithm, denoted as $I_F = \mathsf{FILM}(I_1, I_2)$, elucidating its workflow and design details. 
    \subsection{Workflow Overview}
    Brief and detailed workflows of our FILM paradigm are illustrated in \cref{fig:Workflow,fig:Workflow2}. FILM is segmented into three components: \textit{text feature fusion}, \textit{language-guided vision feature fusion}, and \textit{vision feature decoding}, corresponding to the first, second, and third columns of \cref{fig:Workflow2}, and denoted as $\mathcal{T}(\cdot)$, $\mathcal{V}(\cdot)$, and $\mathcal{D}(\cdot)$, respectively. Specifically, our FILM algorithm takes two source images $\{I_1,I_2\}$ as input, which are initially processed by the \textit{text feature fusion} component $\mathcal{T}$. This component encompasses generating prompts from image caption \cite{DBLP:conf/icml/0008LSH23}, dense caption \cite{nguyen2022grit}, and Segment Anything \cite{Kirillov_2023_ICCV}, to produce textual descriptions via ChatGPT \cite{chatgpt}. The descriptions are encoded via the text encoder of BLIP2 \cite{DBLP:conf/icml/0008LSH23}, and the text features are subsequently fused. The \textit{language-guided vision feature fusion} component $\mathcal{V}$ then utilizes the fused text features to guide the extraction of visual features from the source images using cross-attention. This process identifies and integrates the salient aspects and advantages to be incorporated into the fused image. Finally, the fusion result $F$ is output by the \textit{vision feature decoding} component $\mathcal{D}$, which decodes the fused vision features into an image. Each component's details will be elaborated upon separately.
    
    \subsection{Fusion via Vision-Language Model}
    \bfsection{Component \uppercase\expandafter{\romannumeral1}: Text Feature Fusion}
    In the text feature fusion component, paired source images $\{I_1,I_2\}$ are input, resulting in the fused text feature $\Phi^T_F$, i.e.,
    \begin{equation}\label{}
        \Phi^T_F = \mathcal{T}(I_1,I_2).
    \end{equation}
    Initially, inspired by \citet{DBLP:conf/icml/0008LSH23,nguyen2022grit,Kirillov_2023_ICCV,Image2Paragraph}, we input the images into the BLIP2 \cite{DBLP:conf/icml/0008LSH23}, GRIT \cite{nguyen2022grit}, and Segment Anything \cite{Kirillov_2023_ICCV} models to extract image semantic information from holistic to fine-grained, as \textit{Image Caption}, \textit{Dense Caption}, and \textit{Semantic Mask}. Subsequently, these three prompts are fed into the ChatGPT \cite{chatgpt} model to generate paired text descriptions $\{T_1,T_2\}$ for the source images $\{I_1,I_2\}$. We then input $\{T_1,T_2\}$ into the text encoder of parameter-frozen BLIP2 \cite{DBLP:conf/icml/0008LSH23} model, obtaining the corresponding text features $\{\Phi^T_1, \Phi^T_2\}$. Ultimately, the fused text feature $\Phi^T_F$ is obtained by concatenating $\{\Phi^T_1, \Phi^T_2\}$. For more details on feature prompting and text generation, please refer to \cref{sec:dataset}.
    
    \bfsection{Component \uppercase\expandafter{\romannumeral2}: Language-Guided Vision Feature Fusion}
    In the language-guided vision feature fusion component, we guide the extraction of visual features from the source image through text features, resulting in the visual fusion feature $\Phi^V_F$, i.e.
    \begin{equation}\label{}
        \Phi^V_F = \mathcal{V}(\Phi^T_F,I_1,I_2).
    \end{equation}
    Firstly, source images $\{I_1,I_2\}$ are fed into the image encoders, producing shallow visual features $\{\Phi^{V,(0)}_1, \Phi^{V,(0)}_2\}$ from $\{I_1,I_2\}$, respectively. The image encoder, consisting of Restormer blocks \cite{DBLP:conf/cvpr/ZamirA0HK022} and CNN blocks \cite{HeZRS16}, focuses on both global and local visual representations while maintaining computational efficiency and effective feature extraction. Subsequently, these shallow features are input into the cross-attention mechanism, where fused text features direct the visual feature extraction process, specifically emphasizing aspects of the source image that are desired to be preserved in the output fused image. That is:
    \begin{equation}\label{}
        \Phi^{V,(m)}_1 = \mathcal{CA}\left(\Phi^T_F,\Phi^{V,(m-1)}_1\right),
    \end{equation}
    where $m\!=\!1,\cdots,M$. $\mathcal{CA}(\cdot)$ represents the Cross-Attention module, and $\Phi^{V,(m)}_2$ can be obtained similarly by replacing the subscripts. In $\mathcal{CA}(\cdot)$, \textit{Key}~($K$) and \textit{Value}~($V$) are provided by $\Phi^{V,(m-1)}_1$ or $\Phi^{V,(m-1)}_2$, while \textit{Query}~($Q$) is provided by $\Phi^T_F$.
    Moreover, the feed-forward operation in $\mathcal{CA}(\cdot)$ is also implemented through the Restormer block \cite{DBLP:conf/cvpr/ZamirA0HK022}.
    
    After passing through $M$ Cross-Attention modules, the visual features from text-guided extraction are represented as $\{\Phi^{V,(M)}_1, \Phi^{V,(M)}_2\}$. Subsequently, after the concatenation through channel dimension, $\{\Phi^{V,(M)}_1, \Phi^{V,(M)}_2\}$ yield the fused visual feature $\Phi^V_F$, as shown in \cref{fig:Workflow2}.
    
    \bfsection{Component \uppercase\expandafter{\romannumeral3}: Vision Feature Decoding}
    Finally, the fused visual feature $\Phi^V_F$ is input into the image decoder $\mathcal{D}$, comprising $N$ layers of Restormer \cite{DBLP:conf/cvpr/ZamirA0HK022} and CNN \cite{HeZRS16} blocks, from which the fused image is output, denoted as $I_F = \mathcal{D}(\Phi^V_F)$. $I_F$ refers to the final output fusion image of FILM.
    
    \section{Vision-Language Fusion Dataset}\label{sec:dataset}
    In this section, we will introduce details of the proposed \textit{\textbf{V}ision-\textbf{L}anguage \textbf{F}usion} (\textbf{VLF}) Dataset, including prompts generation, paragraph descriptions output, and representative visualization displays.
    
    \bfsection{Overview} Considering the high computational cost of invoking various vision-language components, and to facilitate subsequent research on image fusion based on vision-language models, we propose the VLF Dataset. This dataset encompasses paired paragraph descriptions generated by ChatGPT, covering all image pairs from the training and test sets of the {eight} widely-used fusion datasets. These include the MSRS \cite{DBLP:journals/inffus/TangYZJM22}, M$^3$FD \cite{DBLP:conf/cvpr/LiuFHWLZL22} and RoadScene \cite{xu2020aaai} datasets for \textit{infrared-visible image fusion} (IVF) task, the Harvard medical dataset \cite{HarvardMedical} for \textit{medical image fusion} (MIF) task, the RealMFF \cite{zhang2020real} and Lytro \cite{nejati2015multi} datasets for \textit{multi-focus image fusion} (MFF) task, and the SICE \cite{cai2018learning} and MEFB \cite{ZHANG2021111} datasets for \textit{multi-exposure image fusion} (MEF) task.
    
    \bfsection{Prompt Generation}
    The output of each component from the Text Paragraph Generation module in FILM is shown in \cref{fig:WorkflowDataset}. Firstly, inspired by \citet{Image2Paragraph}, BLIP2 \cite{DBLP:conf/icml/0008LSH23}, GRIT \cite{nguyen2022grit} and Segment Anything \cite{Kirillov_2023_ICCV} models output {Image Caption}, {Dense Caption}, and {Semantic Mask}, respectively. They provide the one-sentence caption, object-level information, and semantic mask for the input and representing semantic information ranging from coarse-grained to fine-grained.
    
    \bfsection{Generated Paragraph Descriptions} Subsequently, the generated semantic prompts from paired images are input into ChatGPT \cite{chatgpt} to generate paragraph descriptions, which are used to guide subsequent fusion tasks.
    
    \bfsection{Statistical Information}
    This dataset contains 7040 paragraph descriptions, with each description consisting of seven sentences and 186 words on average. We present examples of representative infrared-visible, medical, multi-exposure, multi-focus image pairs in \cref{fig:WorkflowDataset}. More dataset details can be found in \cref{sec:s2}.
    
    \begin{figure*}[t]
        \centering
        \includegraphics[width=\linewidth]{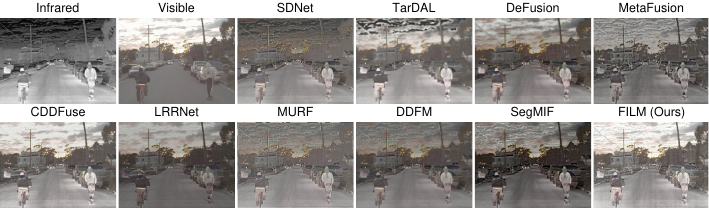}
        \caption{Visualization comparison of the fusion results in the infrared-visible image fusion task.}
        \label{fig:IVF}
        \vspace{-1em}
    \end{figure*}
    
    \begin{table*}[t]
        \centering
        \caption{Quantitative results of IVF. The \colorbox{firstcolor}{red} and \colorbox{secondcolor}{blue} markers represent the best and second-best values, respectively.}
        \label{tab:IVIF}%
        \resizebox{\linewidth}{!}{
            \begin{tabular}{crrrrrrcrrrrrrcrrrrrr}
                \toprule
                \multicolumn{7}{c}{\textbf{MSRS Infrared-Visible Fusion Dataset}} & \multicolumn{7}{c}{\textbf{M$^3$FD Infrared-Visible Fusion Dataset}} & \multicolumn{7}{c}{\textbf{RoadScene Infrared-Visible Fusion Dataset}} \\
                & \multicolumn{1}{c}{EN} & \multicolumn{1}{c}{SD} & \multicolumn{1}{c}{SF} & \multicolumn{1}{c}{AG} & \multicolumn{1}{c}{VIF} & \multicolumn{1}{c}{Qabf} &       & \multicolumn{1}{c}{EN} & \multicolumn{1}{c}{SD} & \multicolumn{1}{c}{SF} & \multicolumn{1}{c}{AG} & \multicolumn{1}{c}{VIF} & \multicolumn{1}{c}{Qabf} &       & \multicolumn{1}{c}{EN} & \multicolumn{1}{c}{SD} & \multicolumn{1}{c}{SF} & \multicolumn{1}{c}{AG} & \multicolumn{1}{c}{VIF} & \multicolumn{1}{c}{Qabf} \\
                \midrule
                SDN   & 5.25  & 17.35 & 8.67  & 2.67  & 0.50  & 0.38  & SDN   & 6.79  & 34.63 & 14.86 & 5.16  & 0.56  & 0.54  & SDN   & 7.34  & 44.74 & 14.99 & 5.94  & 0.62  & 0.55 \\
                TarD  & 5.28  & 25.22 & 5.98  & 1.83  & 0.42  & 0.18  & TarD  & 6.79  & 40.75 & 8.18  & 2.92  & 0.53  & 0.30  & TarD  & 7.25  & 47.57 & 11.46 & 4.23  & 0.56  & 0.43 \\
                DeF   & 6.46  & 37.63 & 8.60  & 2.80  & 0.77  & 0.54  & DeF   & 6.84  & 35.09 & 9.65  & 3.37  & 0.59  & 0.42  & DeF   & 7.39  & 47.60 & 11.26 & 4.47  & 0.63  & 0.50 \\
                Meta  & 5.65  & 24.97 & 9.99  & 3.40  & 0.63  & 0.48  & Meta  & 6.68  & 29.62 & 16.22 & \cellcolor[rgb]{ .945,  .863,  .859}5.68 & 0.68  & 0.57  & Meta  & 6.87  & 31.95 & 14.40 & 5.55  & 0.55  & 0.46 \\
                CDDF  & \cellcolor[rgb]{ .863,  .902,  .941}6.70 & \cellcolor[rgb]{ .945,  .863,  .859}43.39 & \cellcolor[rgb]{ .863,  .902,  .941}11.56 & \cellcolor[rgb]{ .863,  .902,  .941}3.74 & \cellcolor[rgb]{ .863,  .902,  .941}1.05 & \cellcolor[rgb]{ .863,  .902,  .941}0.69 & CDDF  & \cellcolor[rgb]{ .863,  .902,  .941}7.08 & \cellcolor[rgb]{ .863,  .902,  .941}41.29 & \cellcolor[rgb]{ .863,  .902,  .941}16.49 & 5.42  & \cellcolor[rgb]{ .863,  .902,  .941}0.78 & 0.63  & CDDF  & \cellcolor[rgb]{ .863,  .902,  .941}7.41 & \cellcolor[rgb]{ .945,  .863,  .859}54.59 & \cellcolor[rgb]{ .863,  .902,  .941}17.04 & \cellcolor[rgb]{ .863,  .902,  .941}6.07 & 0.63  & 0.51 \\
                LRR   & 6.19  & 31.78 & 8.46  & 2.63  & 0.54  & 0.46  & LRR   & 6.60  & 30.19 & 11.69 & 3.95  & 0.57  & 0.51  & LRR   & 7.09  & 38.77 & 11.50 & 4.36  & 0.43  & 0.33 \\
                MURF  & 5.04  & 20.63 & 10.49 & 3.38  & 0.44  & 0.36  & MURF  & 6.52  & 27.90 & 11.43 & 4.51  & 0.39  & 0.30  & MURF  & 6.91  & 33.46 & 13.74 & 5.31  & 0.53  & 0.47 \\
                DDFM  & 6.19  & 29.26 & 7.44  & 2.51  & 0.73  & 0.48  & DDFM  & 6.72  & 31.15 & 9.84  & 3.42  & 0.63  & 0.47  & DDFM  & 7.27  & 42.94 & 10.89 & 4.20  & 0.63  & 0.50 \\
                SegM  & 5.95  & 37.28 & 11.10 & 3.47  & 0.88  & 0.63  & SegM  & 6.89  & 35.64 & 16.11 & 5.52  & 0.78  & \cellcolor[rgb]{ .863,  .902,  .941}0.65 & SegM  & 7.29  & 47.10 & 15.07 & 5.78  & \cellcolor[rgb]{ .863,  .902,  .941}0.65 & \cellcolor[rgb]{ .863,  .902,  .941}0.56 \\
                Ours  & \cellcolor[rgb]{ .945,  .863,  .859}6.72 & \cellcolor[rgb]{ .863,  .902,  .941}43.17 & \cellcolor[rgb]{ .945,  .863,  .859}11.70 & \cellcolor[rgb]{ .945,  .863,  .859}3.84 & \cellcolor[rgb]{ .945,  .863,  .859}1.06 & \cellcolor[rgb]{ .945,  .863,  .859}0.73 & Ours  & \cellcolor[rgb]{ .945,  .863,  .859}7.09 & \cellcolor[rgb]{ .945,  .863,  .859}41.53 & \cellcolor[rgb]{ .945,  .863,  .859}16.77 & \cellcolor[rgb]{ .863,  .902,  .941}5.55 & \cellcolor[rgb]{ .945,  .863,  .859}0.83 & \cellcolor[rgb]{ .945,  .863,  .859}0.67 & Ours  & \cellcolor[rgb]{ .945,  .863,  .859}7.43 & \cellcolor[rgb]{ .863,  .902,  .941}49.25 & \cellcolor[rgb]{ .945,  .863,  .859}17.34 & \cellcolor[rgb]{ .945,  .863,  .859}6.60 & \cellcolor[rgb]{ .945,  .863,  .859}0.69 & \cellcolor[rgb]{ .945,  .863,  .859}0.62 \\
                \bottomrule
        \end{tabular}}
        \vspace{-1em}
    \end{table*}%

    \begin{figure}[t]
        \centering
        \includegraphics[width=\linewidth]{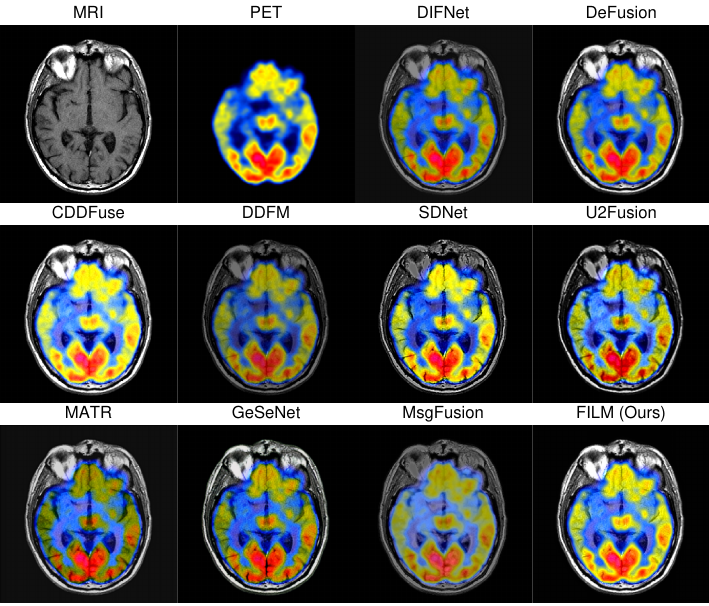}
        \caption{Visualization comparison of the fusion results in the medical image fusion task.}
        \label{fig:MIF}
        \vspace{-1em}
    \end{figure}
    
    \begin{table}[t]
        \centering
        \caption{Quantitative results of MIF. The \colorbox{firstcolor}{red} and \colorbox{secondcolor}{blue} markers represent the best and second-best values, respectively.}
        \label{tab:MIF}%
        \resizebox{0.95\linewidth}{!}{
            \begin{tabular}{crrrrrr}
                \toprule
                \multicolumn{7}{c}{\textbf{Harvard Medical Image Fusion Dataset}} \\
                & \multicolumn{1}{c}{EN} & \multicolumn{1}{c}{SD} & \multicolumn{1}{c}{SF} & \multicolumn{1}{c}{AG} & \multicolumn{1}{c}{VIF} & \multicolumn{1}{c}{Qabf} \\
                \midrule
                DIFNet & \cellcolor[rgb]{ .863,  .902,  .941}4.58 & 49.99 & 14.93 & 4.09  & 0.61  & 0.59 \\
                DeFusion & 3.90  & 54.77 & 16.87 & 4.30  & 0.62  & 0.57 \\
                CDDFuse & 4.00  & \cellcolor[rgb]{ .863,  .902,  .941}70.58 & 22.84 & 5.75  & 0.71  & 0.69 \\
                DDFM  & 3.82  & 56.47 & 16.17 & 4.16  & 0.68  & 0.65 \\\cdashline{1-7}[1pt/1pt]
                SDNet & 3.53  & 48.85 & \cellcolor[rgb]{ .863,  .902,  .941}23.15 & 5.53  & 0.54  & 0.63 \\
                U2Fusion & 3.56  & 49.95 & 19.70 & 4.98  & 0.47  & 0.53 \\
                MATR  & 4.09  & 48.63 & 17.87 & 4.70  & 0.75  & 0.72 \\
                GeSeNet & 4.31  & 62.47 & 22.72 & \cellcolor[rgb]{ .863,  .902,  .941}5.85 & \cellcolor[rgb]{ .863,  .902,  .941}0.76 & \cellcolor[rgb]{ .863,  .902,  .941}0.76 \\
                MsgFusion & 4.06  & \cellcolor[rgb]{ .945,  .863,  .859}75.01 & 20.34 & 5.09  & 0.49  & 0.50 \\
                Ours  & \cellcolor[rgb]{ .945,  .863,  .859}4.74 & 65.26 & \cellcolor[rgb]{ .945,  .863,  .859}23.36 & \cellcolor[rgb]{ .945,  .863,  .859}6.19 & \cellcolor[rgb]{ .945,  .863,  .859}0.78 & \cellcolor[rgb]{ .945,  .863,  .859}0.76 \\
                \bottomrule
        \end{tabular}}
        %\vspace{-2em}
    \end{table}%
    
    \begin{figure*}[t]
        \centering
        \includegraphics[width=\linewidth]{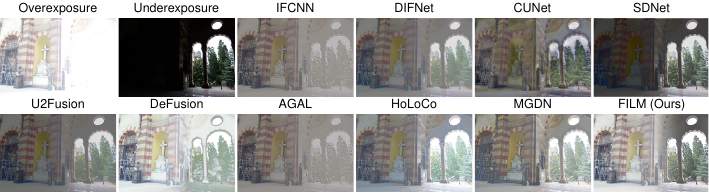}
        \caption{Visualization comparison of the fusion results in the multi-exposure image fusion task.}
        \label{fig:MEF}
        %\vspace{-1em}
    \end{figure*}
    
    \begin{table*}[t]
        \centering
        \caption{Quantitative results of MEF. The \colorbox{firstcolor}{red} and \colorbox{secondcolor}{blue} markers represent the best and second-best values, respectively.}
        \label{tab:MEF}%
        \resizebox{0.9\linewidth}{!}{
            \begin{tabular}{crrrrrrcrrrrrr}
                \toprule
                \multicolumn{7}{c}{\textbf{SICE Multi-exposure Image Fusion Dataset}} & \multicolumn{7}{c}{\textbf{MEFB Multi-exposure Image Fusion Dataset}} \\
                & \multicolumn{1}{c}{EN} & \multicolumn{1}{c}{SD} & \multicolumn{1}{c}{SF} & \multicolumn{1}{c}{AG} & \multicolumn{1}{c}{VIF} & \multicolumn{1}{c}{Qabf} &       & \multicolumn{1}{c}{EN} & \multicolumn{1}{c}{SD} & \multicolumn{1}{c}{SF} & \multicolumn{1}{c}{AG} & \multicolumn{1}{c}{VIF} & \multicolumn{1}{c}{Qabf} \\
                \midrule
                IFCNN & 6.67  & 39.43 & 16.93 & 4.59  & 0.73  & \cellcolor[rgb]{ .863,  .902,  .941}0.71 & IFCNN & 6.99  & 52.49 & 18.16 & 5.34  & 0.71  & \cellcolor[rgb]{ .863,  .902,  .941}0.69 \\
                DIFNet & 6.56  & 35.76 & 11.86 & 3.09  & 0.46  & 0.50  & DIFNet & 6.99  & 50.23 & 11.79 & 3.47  & 0.51  & 0.53 \\
                CUNet & 6.90  & 34.18 & 11.87 & 3.80  & 0.69  & 0.50  & CUNet & 7.18  & 45.37 & 12.78 & 4.28  & 0.71  & 0.50 \\
                SDNet & 6.47  & 38.25 & \cellcolor[rgb]{ .863,  .902,  .941}19.34 & 4.80  & 0.48  & 0.45  & SDNet & 6.59  & 51.77 & \cellcolor[rgb]{ .863,  .902,  .941}20.53 & 5.27  & 0.55  & 0.42 \\
                U2Fusion & 6.43  & 34.77 & 10.71 & 3.17  & 0.48  & 0.57  & U2Fusion & 6.67  & 46.73 & 12.54 & 3.82  & 0.51  & 0.56 \\
                DeFusion & 6.87  & 44.73 & 14.28 & 4.04  & 0.87  & 0.57  & DeFusion & 7.10  & 56.46 & 14.86 & 4.48  & 0.70  & 0.59 \\
                AGAL  & \cellcolor[rgb]{ .863,  .902,  .941}7.06 & \cellcolor[rgb]{ .863,  .902,  .941}46.03 & 16.64 & \cellcolor[rgb]{ .863,  .902,  .941}4.91 & 0.72  & 0.53  & AGAL  & 7.14  & \cellcolor[rgb]{ .863,  .902,  .941}60.63 & 17.77 & 5.33  & 0.79  & 0.65 \\
                HoLoCo & 7.04  & 42.73 & 9.33  & 3.47  & 0.74  & 0.37  & HoLoCo & 7.20  & 53.88 & 12.80 & 4.34  & 0.73  & 0.54 \\
                MGDN  & 6.94  & 43.69 & 15.04 & 4.59  & \cellcolor[rgb]{ .863,  .902,  .941}0.88 & 0.64  & MGDN  & \cellcolor[rgb]{ .863,  .902,  .941}7.25 & 55.97 & 18.09 & \cellcolor[rgb]{ .863,  .902,  .941}5.76 & \cellcolor[rgb]{ .863,  .902,  .941}0.96 & 0.65 \\
                Ours  & \cellcolor[rgb]{ .945,  .863,  .859}7.07 & \cellcolor[rgb]{ .945,  .863,  .859}54.21 & \cellcolor[rgb]{ .945,  .863,  .859}19.42 & \cellcolor[rgb]{ .945,  .863,  .859}5.15 & \cellcolor[rgb]{ .945,  .863,  .859}1.05 & \cellcolor[rgb]{ .945,  .863,  .859}0.79 & Ours  & \cellcolor[rgb]{ .945,  .863,  .859}7.31 & \cellcolor[rgb]{ .945,  .863,  .859}69.02 & \cellcolor[rgb]{ .945,  .863,  .859}20.98 & \cellcolor[rgb]{ .945,  .863,  .859}6.15 & \cellcolor[rgb]{ .945,  .863,  .859}0.98 & \cellcolor[rgb]{ .945,  .863,  .859}0.77 \\
                \bottomrule
        \end{tabular}}
        %\vspace{-1.5em}
    \end{table*}%

    \begin{figure*}[t]
        \centering
        \includegraphics[width=0.8\linewidth]{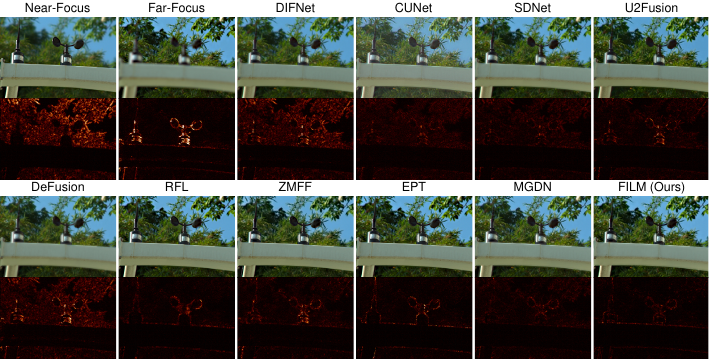}
        \caption{Visualization comparison of the fusion results and error maps in the multi-focus image fusion task.}
        \label{fig:MFF}
        %\vspace{-1em}
    \end{figure*}
    
    \begin{table*}[t]
        \centering
        \caption{Quantitative results of MFF. The \colorbox{firstcolor}{red} and \colorbox{secondcolor}{blue} markers represent the best and second-best values, respectively.}
        \label{tab:MFF}%
        \resizebox{0.9\linewidth}{!}{
            \begin{tabular}{crrrrrrcrrrrrr}
                \toprule
                \multicolumn{7}{c}{\textbf{RealMFF Multi-focus Image Fusion Dataset}} & \multicolumn{7}{c}{\textbf{Lytro Multi-focus Image Fusion on Dataset}} \\
                & \multicolumn{1}{c}{EN} & \multicolumn{1}{c}{SD} & \multicolumn{1}{c}{SF} & \multicolumn{1}{c}{AG} & \multicolumn{1}{c}{VIF} & \multicolumn{1}{c}{Qabf} &       & \multicolumn{1}{c}{EN} & \multicolumn{1}{c}{SD} & \multicolumn{1}{c}{SF} & \multicolumn{1}{c}{AG} & \multicolumn{1}{c}{VIF} & \multicolumn{1}{c}{Qabf} \\
                \midrule
                DIFNet & 7.01  & 51.17 & 10.78 & 3.96  & 0.89  & 0.69  & DIFNet & 7.43  & 52.52 & 11.47 & 4.30  & 0.73  & 0.54 \\
                CUNet & 6.72  & 38.97 & 13.59 & 4.81  & 0.77  & 0.65  & CUNet & 7.25  & 45.78 & 15.54 & 5.58  & 0.71  & 0.65 \\
                SDNet & 6.95  & 50.96 & \cellcolor[rgb]{ .863,  .902,  .941}15.22 & 5.02  & 0.93  & 0.73  & SDNet & 7.47  & 55.25 & 16.88 & 5.84  & 0.84  & 0.69 \\
                U2Fusion & 6.77  & 48.49 & 14.07 & 5.09  & 0.95  & 0.70  & U2Fusion & 7.30  & 51.95 & 14.83 & 5.60  & 0.83  & 0.65 \\
                DeFusion & 7.09  & \cellcolor[rgb]{ .863,  .902,  .941}54.42 & 11.24 & 4.08  & 0.98  & 0.69  & DeFusion & 7.52  & 56.65 & 11.55 & 4.35  & 0.80  & 0.55 \\
                RFL   & 7.00  & 51.62 & 14.93 & 5.03  & 0.96  & 0.75  & RFL   & 7.53  & 57.53 & 18.43 & \cellcolor[rgb]{ .863,  .902,  .941}6.84 & 0.94  & 0.73 \\
                ZMFF  & 6.99  & 51.15 & 13.93 & 4.95  & 0.94  & 0.70  & ZMFF  & 7.53  & 56.96 & \cellcolor[rgb]{ .863,  .902,  .941}18.84 & 6.76  & 0.93  & 0.69 \\
                EPT   & 7.00  & 51.64 & 14.97 & 5.04  & 0.96  & \cellcolor[rgb]{ .863,  .902,  .941}0.75 & EPT   & 7.53  & \cellcolor[rgb]{ .863,  .902,  .941}57.55 & 18.44 & 6.84  & \cellcolor[rgb]{ .863,  .902,  .941}0.94 & \cellcolor[rgb]{ .863,  .902,  .941}0.74 \\
                MGDN  & \cellcolor[rgb]{ .863,  .902,  .941}7.09 & 54.24 & 15.15 & \cellcolor[rgb]{ .863,  .902,  .941}5.24 & \cellcolor[rgb]{ .863,  .902,  .941}1.07 & 0.75  & MGDN  & \cellcolor[rgb]{ .863,  .902,  .941}7.54 & 57.50 & 18.81 & 6.67  & 0.93  & 0.74 \\
                Ours  & \cellcolor[rgb]{ .945,  .863,  .859}7.11 & \cellcolor[rgb]{ .945,  .863,  .859}54.93 & \cellcolor[rgb]{ .945,  .863,  .859}15.62 & \cellcolor[rgb]{ .945,  .863,  .859}5.43 & \cellcolor[rgb]{ .945,  .863,  .859}1.10 & \cellcolor[rgb]{ .945,  .863,  .859}0.76 & Ours  & \cellcolor[rgb]{ .945,  .863,  .859}7.56 & \cellcolor[rgb]{ .945,  .863,  .859}59.15 & \cellcolor[rgb]{ .945,  .863,  .859}19.57 & \cellcolor[rgb]{ .945,  .863,  .859}6.97 & \cellcolor[rgb]{ .945,  .863,  .859}0.98 & \cellcolor[rgb]{ .945,  .863,  .859}0.74 \\
                \bottomrule
        \end{tabular}}
        %\vspace{-1.5em}
    \end{table*}%
    
    \begin{table*}[t]
        \centering
        \caption{Ablation experiments results, with \colorbox{firstcolor}{red} resent best values.}
        \label{tab:ablation}%
        \resizebox{\linewidth}{!}{
            \begin{tabular}{lcccccccccc}
                \toprule
                \multicolumn{1}{c}{\multirow{2}{*}{Descriptions}} & \multicolumn{4}{c}{Configurations}   & \multicolumn{6}{c}{Metrics}     \\
                \cmidrule(lr){2-5}\cmidrule(lr){6-11}
                & {Image caption} & {Dense caption} & {Segment mask} & {GPT} & {EN} & {SD} & {SF} & {AG} & {VIF} & {Qabf} \\
                \midrule
                Exp.~\uppercase\expandafter{\romannumeral1}: w/o text &       &       &       &       & 7.16  & 43.45 & 11.58 & 5.63  & 0.51  & 0.48 \\
                Exp.~\uppercase\expandafter{\romannumeral2}: w/o caption &       &       &       & \Checkmark     & 7.25  & 45.66 & 11.94 & 5.99  & 0.54  & 0.51 \\
                Exp.~\uppercase\expandafter{\romannumeral3}: w/o DC and SM & \Checkmark     &       &       & \Checkmark     & 7.26  & 47.71 & 11.87 & 6.05  & 0.55  & 0.53 \\
                Exp.~\uppercase\expandafter{\romannumeral4}: w/o SM & \Checkmark     & \Checkmark     &       & \Checkmark     & 7.33  & 49.09 & 16.36 & 6.94  & 0.62  & 0.55 \\
                Exp.~\uppercase\expandafter{\romannumeral5}: w/o GPT & \Checkmark     & \Checkmark     & \Checkmark     &       & 7.29  & 50.38 & 14.39 & 6.55  & 0.58  & 0.53 \\
                \midrule
                FILM (Ours)  & \Checkmark     & \Checkmark     & \Checkmark     & \Checkmark     & \cellcolor[rgb]{ .949,  .863,  .859}7.43 & \cellcolor[rgb]{ .949,  .863,  .859}49.25 & \cellcolor[rgb]{ .949,  .863,  .859}17.34 & \cellcolor[rgb]{ .949,  .863,  .859}6.60 & \cellcolor[rgb]{ .949,  .863,  .859}0.69 & \cellcolor[rgb]{ .949,  .863,  .859}0.62 \\
                \bottomrule
        \end{tabular}}
    \end{table*}%

    \section{Experiment}
    In this section, we will demonstrate the performance of FILM on various image fusion tasks, showcasing its superiority. Due to space constraints, more visual results are presented in the supplementary material (\cref{sec:s3}).
    
    \bfsection{Loss Function}
    For the total training loss, we set it as:
    \begin{equation}\label{equ:loss1}
        \mathcal{L}_{total}=\mathcal{L}_{int} + \alpha_1\mathcal{L}_{grad} +\alpha_2\mathcal{L}_{\text{SSIM}},
    \end{equation}
    where $\alpha_1$, $\alpha_2$ are tuning parameters.
    In the IVF task, following the setting in \citet{Zhao_2023_CVPR}, $\mathcal{L}_{int}\!=\!\frac{1}{HW}\!\|I_F\!-\!\max (I_1, I_2)\|_1$, and  $\mathcal{L}_{grad}\!=\!\frac{1}{HW}\!\|\left|\nabla I_F\right|\!-\!\max (\left|\nabla I_{1}\right|\!,\!\left|\nabla I_{2}\right|)\|_1$. $\nabla$ indicates the Sobel gradient operator.
    $\alpha_1$ and $\alpha_2$ are set to 20 and 0, respectively.
    MIF task does not need fine-tuning training, therefore it has no loss function.
    For MFF and MEF tasks, inspired by \citet{liu2023holoco}, we set $\mathcal{L}_{int}\!=\!\frac{1}{HW}\!\|I_F\!- mean (I_{1}, I_{2})\|_1$, $\mathcal{L}_{grad}\!=\!\frac{1}{HW}\!\|\left|\nabla I_F\right|\!-\!\max (\left|\nabla I_{1}\right|\!,\!\left|\nabla I_{2}\right|)\|_1$, and $\mathcal{L}_{\text{SSIM}} = 2 - \text{SSIM}(I_1, I_F) - \text{SSIM}(I_2,I_F)$. $\{\alpha_1,\alpha_2\}$ are set to $\{300,1\}$ and $\{500,1\}$ in MFF and MEF tasks respectively, in order to ensure the magnitude comparable in each term.
    
    \bfsection{Training Details}
    A machine with {eight} NVIDIA GeForce RTX 3090 GPUs is utilized for our experiments. We train the network for 300 epochs using the Adam optimizer, with an initial learning rate of 1e-4 and decreasing by a factor of 0.5 every 50 epochs. The Adam optimization strategy is employed with the batchsize set as 16.
    We incorporate Restormer blocks \cite{DBLP:conf/cvpr/ZamirA0HK022} in both language-guided vision encoder $\mathcal{V}(\cdot)$ and vision feature decoder $\mathcal{D}(\cdot)$, with each block having 8 attention heads and a dimensionality of 64. $M$ and $N$, representing the number of blocks in $\mathcal{V}(\cdot)$ and $\mathcal{D}(\cdot)$, are set to 2 and 3, respectively.
    
    \bfsection{Metrics}
    We employ six quantitative metrics to assess the fusion outcomes: entropy (EN), standard deviation (SD), spatial frequency (SF), average gradient (AG), visual information fidelity (VIF) and $Q^{AB/F}$. Higher metric values indicate superior quality in the fused image. Further information is available in \citet{ma2019infrared}.
    
    \subsection{Infrared and Visible Image Fusion}
    \bfsection{Setup}
    Following \citet{Zhao_2023_CVPR,Zhao_2023_ICCV}, infrared-visible fusion experiments are conducted on the MSRS \cite{DBLP:journals/inffus/TangYZJM22}, M$^3$FD \cite{DBLP:conf/cvpr/LiuFHWLZL22} and RoadScene \cite{xu2020aaai} datasets. 1083 image pairs in MSRS are for training and 361 pairs are for testing. The generalizability of FILM is further assessed by M$^3$FD and RoadScene without finetuning. We evaluated FILM against various state-of-the-art (SOTA) methods including SDNet \cite{DBLP:journals/ijcv/ZhangM21}, TarDAL \cite{DBLP:conf/cvpr/LiuFHWLZL22}, DeFusion \cite{Liang2022ECCV}, MetaFusion \cite{DBLP:conf/cvpr/ZhaoXZHL23}, CDDFuse \cite{Zhao_2023_CVPR}, LRRNet \cite{li2023lrrnet}, MURF \cite{xu2023murf}, DDFM \cite{Zhao_2023_ICCV}, and SegMIF \cite{Liu_2023_ICCV}.
    
    \bfsection{Comparison with SOTA Methods}
    In \cref{fig:IVF}, FILM successfully integrated the thermal radiation information with the detailed texture features. Leveraging textual features and knowledge, the fusion process enhanced the visibility of objects in low-light environments, making textures and contours clearer, and reducing artifacts. For the quantitative results in \cref{tab:IVIF}, our method showcases exceptional performance in almost all metrics, confirming its adaptability for various environmental scenarios and object categories. Hence, FILM is proven to well maintain the completeness and richness of the information from source images, and generate results that conform to human visual perception.
    
    \subsection{Medical Image Fusion}
    \bfsection{Setup}
    Following \citet{Zhao_2023_ICCV}, we engage the Harvard Medical dataset \cite{HarvardMedical}, which consisted of 50 pairs of MRI-CT, MRI-PET, and MRI-SPECT images, to evaluate the generalizability of our model. Notably, we employ the model trained on the IVF experiments and conducted a generalization test on the Harvard Medical dataset without any fine-tuning. The competitors include
    DIFNet \cite{jung2020unsupervised}, SDNet \cite{DBLP:journals/ijcv/ZhangM21}, U2Fusion \cite{9151265}, DeFusion \cite{Liang2022ECCV}, MATR \cite{tang2022matr}, CDDFuse \cite{Zhao_2023_CVPR}, DDFM \cite{Zhao_2023_ICCV}, GeSeNet \cite{li2023gesenet} and MsgFusion \cite{wen2023msgfusion}.
    Results from DIFNet, DeFusion, CDDFuse, and DDFM are the generalized outcomes of IVF models without fine-tuning, whereas the other results are from models specialized training using the MIF datasets.
    
    \bfsection{Comparison with SOTA Methods}
    In terms of visual perception and quantitative analysis (\cref{fig:MIF} and \cref{tab:MIF}), FILM has shown outstanding accuracy in extracting cross-modal structural highlights and detailed texture features, effectively integrating source information into the fused images. These achievements surpass even those of fusion models specifically fine-tuned via medical image pairs.
    
    \subsection{Multi-exposure Image Fusion}
    \bfsection{Setup}
    We conduct MEF experiments on the SICE \cite{cai2018learning} and MEFB \cite{ZHANG2021111} dataset. We utilized 499 pairs from SICE dataset for training, while 90 pairs from SICE and 40 pairs from MEFB for testing. Our comparison methods encompass IFCNN \cite{DBLP:journals/inffus/ZhangLSYZZ20}, DIFNet \cite{jung2020unsupervised}, CUNet \cite{DBLP:journals/pami/0002D21}, SDNet \cite{DBLP:journals/ijcv/ZhangM21}, U2Fusion \cite{9151265}, DeFusion \cite{Liang2022ECCV}, AGAL \cite{liu2022attention},  HoLoCo \cite{liu2023holoco} and MGDN \cite{guan2023mutual}.
    
    \bfsection{Comparison with SOTA Methods}
    Both quantitative and qualitative results in \cref{tab:MEF} and \cref{fig:MEF} demonstrate the effectiveness of FILM, which adeptly handles multiple images with varying exposures, expanding the dynamic range while simultaneously improving image quality and enhancing contrast.
    
    \subsection{Multi-focus Image Fusion}
    \bfsection{Setup}
    MFF experiments are conducted using RealMFF \cite{zhang2020real} and Lytro \cite{nejati2015multi}. 639 image pairs from RealMFF are employed for training, while 71 pairs from it are reserved for testing and 20 image pairs in Lytro are utilized for generalizability test. Comparative methods encompass DIFNet \cite{jung2020unsupervised}, CUNet \cite{DBLP:journals/pami/0002D21}, SDNet \cite{DBLP:journals/ijcv/ZhangM21}, U2Fusion \cite{9151265}, DeFusion \cite{Liang2022ECCV}, RFL \cite{wang2022self}, ZMFF \cite{hu2023zmff}, EPT \cite{wang2023multi}, and MGDN \cite{guan2023mutual}.
    
    \bfsection{Comparison with SOTA Methods}
    As illustrated in \cref{fig:MFF}, benefiting from textual descriptions, FILM excels in identifying clear regions within multi-focus image pairs, ensuring sharp foreground and background elements. The quantitative results in \cref{tab:MFF} further underscore the excellence of our methodology.
    
    \subsection{Ablation Studies}
    To explore the effectiveness of each module in our proposed method, using the infrared-visible fusion task as an example, we conduct ablation studies on the test dataset of RoadScene \cite{xu2020aaai}. The results are presented in \cref{tab:ablation}.
    
    \bfsection{Textual Guidance}
    In Exp.~\uppercase\expandafter{\romannumeral1}, we remove the guidance through textual information and only use image features for fusion, i.e., the cross-attention layers between text and image features are eliminated, aiming to demonstrate the effect of text-guided feature extraction and fusion in FILM. By increasing the number of Restormer blocks, we maintain the total number of parameters close to the original model.
    
    \bfsection{Semantic Prompts}
    Then, in Exp.~\uppercase\expandafter{\romannumeral2}-\uppercase\expandafter{\romannumeral4}, we test the guiding role of text semantic prompts from holistic to fine-grained, including image caption (IC), dense caption (DC), and segment mask (SM).
    In Exp.~\uppercase\expandafter{\romannumeral2}, we directly feed the source images into ChatGPT. By manually providing prompts, GPT generates overall descriptions of the images, which are used as text inputs for image fusion. This study bypassed the steps involving prompts from IC, DC and SM.
    In Exp.~\uppercase\expandafter{\romannumeral3}, only IC is input into GPT, whereas in Exp.~\uppercase\expandafter{\romannumeral4}, both IC and DC are together input into GPT, revealing the importance of different aspects of the captions from coarse-grained to fine-grained.
    
    \bfsection{ChatGPT}
    Finally, in Exp.~\uppercase\expandafter{\romannumeral5}, after extracting IC, DC and SM from images, we directly concatenate these three captions as the text description without inputting them into GPT. This is to demonstrate GPT's capability in integrating textual information and its effort for fusion performance.
    
    In conclusion, ablation experiments demonstrate that relying on the comprehensive information from different grains of captions and the powerful summarization capability of GPT, our experimental setup achieved optimal fusion performance, validating the rationality of our FILM setting.
    
    \section{Conclusion}
    
    This study addresses a significant shortcoming of existing image fusion techniques: their insufficient exploitation of deeper semantic information beyond visual features. To this end, we present, for the first time, a novel paradigm called \textit{Image \textbf{F}usion via V\textbf{I}sion-\textbf{L}anguage \textbf{M}odel} (\textbf{FILM}), which employs explicit textural descriptions of source images from large language models to guide and enhance the fusion process, enabling a more comprehensive understanding of image content.     
    Furthermore, we explore the feasibility of integrating the vision-language model framework into the image fusion process. Notably, in FILM, any component within the model, such as BLIP2 or ChatGPT, is replaceable.
    
    FILM has shown promising results on various image fusion tasks, including infrared-visible, medical, multi-exposure and multi-focus scenarios. In addition, we present a novel benchmark vision-language dataset, including ChatGPT-generated descriptions for eight image fusion datasets. We hope that our study will open up new opportunities for large-scale vision-language models in the realm of image fusion.
    
    \section*{Acknowledgements}
    This work has been supported by the National Natural Science Foundation of China under Grant 12371512, Shanghai Municipal Science and Technology Major Project under Grant 2021SHZDZX0102 and the Fundamental Research Funds for the Central Universities.
    
    \section*{Impact Statement}
    This paper presents work whose goal is to advance the field of Machine Learning. There are many potential societal consequences of our work, none of which we feel must be specifically highlighted here.
    
    \bibliography{xbib}
    \bibliographystyle{icml2024}

    \appendix
    \onecolumn
    
    \section{More Visualization Results for VLF dataset}\label{sec:s2}
\begin{figure*}[t]
        \centering
        \includegraphics[width=\linewidth]{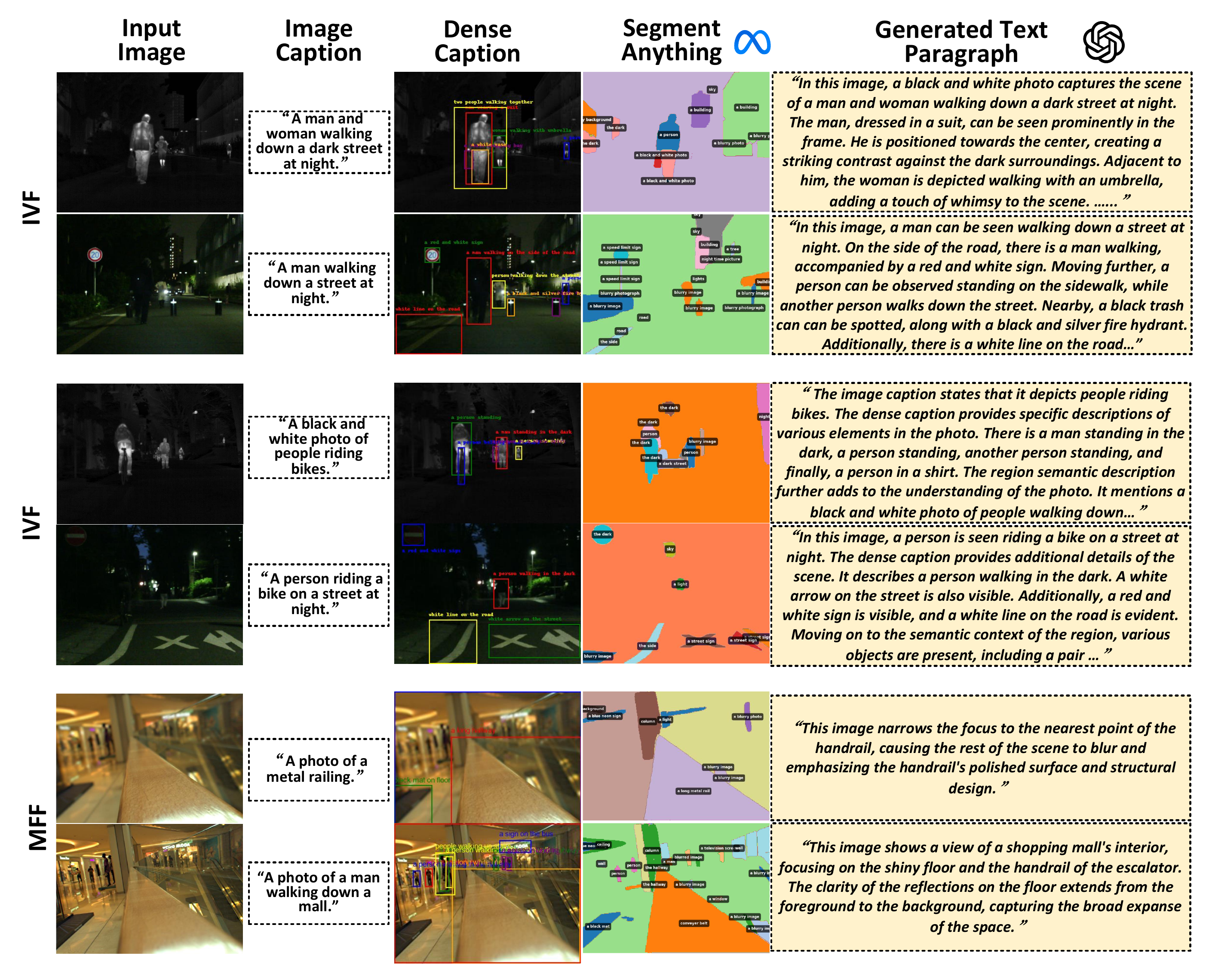}
        \caption{More visualization results for the VLF dataset on IVF and MFF.}
        \label{fig:VLFs1}
    \end{figure*}

\begin{figure*}[t]
        \centering
        \includegraphics[width=\linewidth]{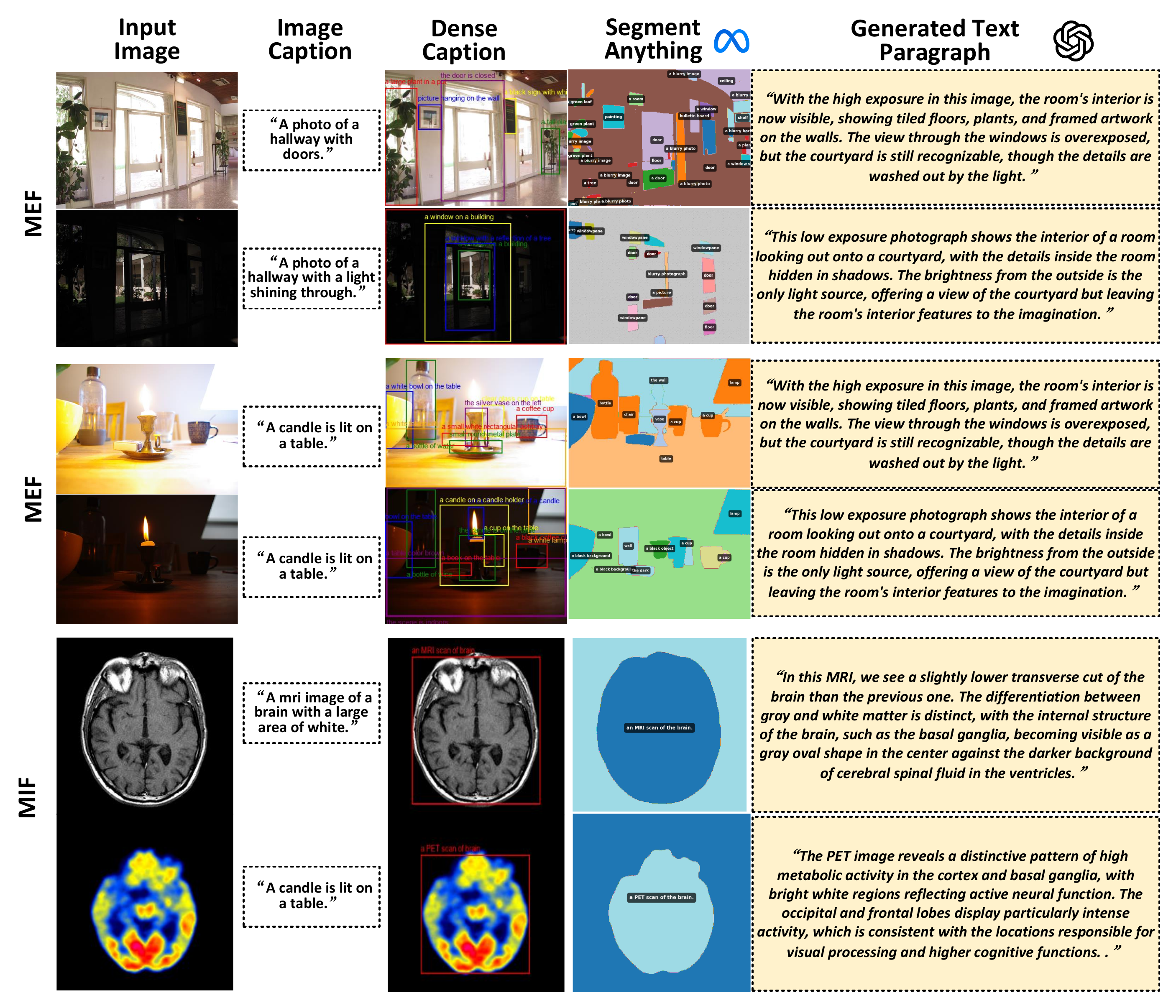}
        \caption{More visualization results for the VLF dataset on MEF and MIF.}
        \label{fig:VLFs2}
    \end{figure*}

    More visualization results for the VLF dataset are displayed in \cref{fig:VLFs1,fig:VLFs2}.    
    
    \section{Detailed Illustration to Datasets}\label{sec:s1}
    We adopt widely-used benchmarks MSRS~\cite{DBLP:journals/inffus/TangYZJM22}, M$^3$FD~\cite{DBLP:conf/cvpr/LiuFHWLZL22} and  RoadScene~\cite{xu2020aaai} for \textit{Infrared-Visible image Fusion}~(IVF), Harvard Medical dataset \cite{HarvardMedical} for \textit{Medical Image Fusion}~(MIF), SICE \cite{cai2018learning} and MEFB \cite{ZHANG2021111} dataset for \textit{Multi-exposure Image Fusion}~(MEF), as well as RealMFF \cite{zhang2020real} and Lytro \cite{nejati2015multi} dataset for \textit{Multi-focus Image Fusion}~(MFF), respectively.
    \begin{itemize}
        \item MSRS dataset\footnote{\href{https://github.com/Linfeng-Tang/MSRS}{https://github.com/Linfeng-Tang/MSRS}}: 1083 pairs for IVF training and 361 pairs for IVF testing.
        \item M$^3$FD dataset\footnote{\href{https://github.com/JinyuanLiu-CV/TarDAL}{https://github.com/JinyuanLiu-CV/TarDAL}}: {100} pairs for IVF testing.
        \item RoadScene dataset\footnote{\href{https://github.com/hanna-xu/RoadScene}{https://github.com/hanna-xu/RoadScene} }: {70} pairs for IVF validation, {70} pairs for IVF testing.
        \item Harvard Medical Image dataset\footnote{\href{http://www.med.harvard.edu/AANLIB/home.html}{http://www.med.harvard.edu/AANLIB/home.html}}: {50} pairs for MIF testing.
        \item SICE dataset\footnote{\href{https://github.com/csjcai/SICE}{https://github.com/csjcai/SICE}}: 499 pairs for MEF training and 90 pairs MEF testing.
        \item MEFB dataset\footnote{\href{https://github.com/xingchenzhang/MEFB}{https://github.com/xingchenzhang/MEFB}}: 40 pairs for MEF testing.
        \item RealMFF dataset\footnote{\href{https://github.com/Zancelot/Real-MFF}{https://github.com/Zancelot/Real-MFF}}: 639 pairs for MFF training and 71 pairs for MFF testing.
        \item Lytro dataset\footnote{\href{http://mansournejati.ece.iut.ac.ir/content/lytro-multi-focus-dataset}{http://mansournejati.ece.iut.ac.ir/content/lytro-multi-focus-dataset}}: 20 pairs for testing.
    \end{itemize}


    \section{More Qualitative Comparison Fusion Results}\label{sec:s3}
\begin{figure*}[t]
        \centering
        \includegraphics[width=\linewidth]{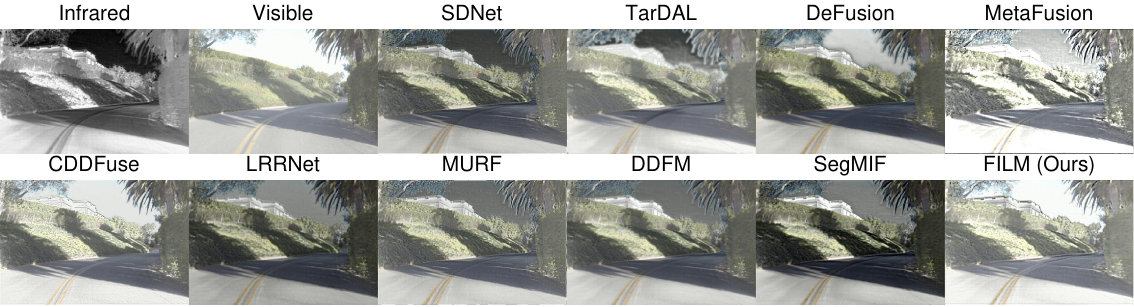}
        \caption{Visualization comparison of the fusion results in the infrared-visible image fusion task.}
        \label{fig:IVF1}
    \end{figure*}

\begin{figure}[t]
        \centering
        \includegraphics[width=\linewidth]{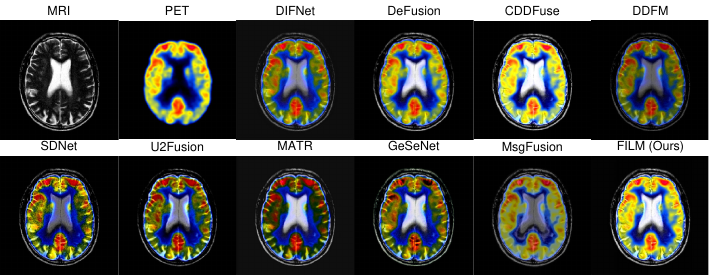}
        \caption{Visualization comparison of the fusion results in the medical image fusion task.}
        \label{fig:MIF1}
    \end{figure}

\begin{figure*}[t]
        \centering
        \includegraphics[width=\linewidth]{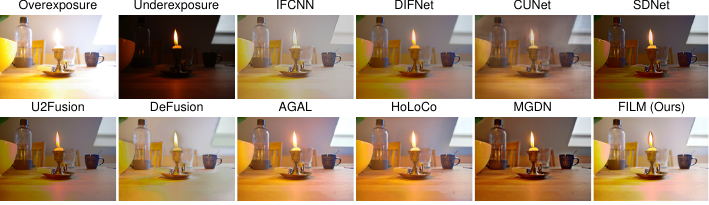}
        \caption{Visualization comparison of the fusion results in the multi-exposure image fusion task.}
        \label{fig:MEF1}
    \end{figure*}

\begin{figure*}[t]
        \centering
        \includegraphics[width=\linewidth]{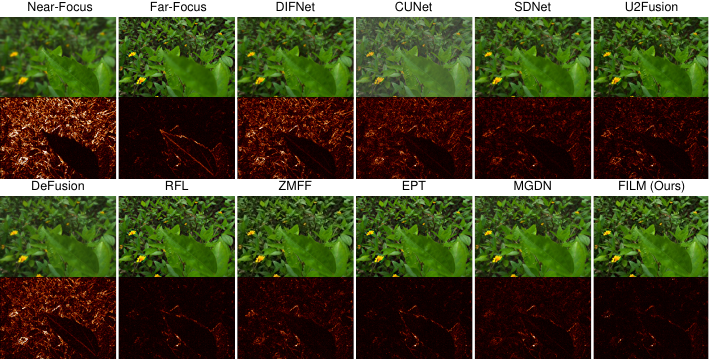}
        \caption{Visualization comparison of the fusion results and error maps in the multi-focus image fusion task.}
        \label{fig:MFF1}
    \end{figure*}

    \begin{itemize}
        \item More qualitative comparisons for \textit{Infrared-Visible image Fusion} are shown in \cref{fig:IVF1}.
        \item More qualitative comparisons for \textit{Medical Image Fusion} are shown in \cref{fig:MIF1}.
        \item More qualitative comparisons for \textit{Multi-exposure Image Fusion} are shown in \cref{fig:MEF1}.
        \item More qualitative comparisons for \textit{Multi-focus Image Fusion} are shown in \cref{fig:MFF1}.
    \end{itemize}

\end{document}